\documentclass{article}
\usepackage{ijcai24}

\usepackage{times}
\usepackage{soul}
\usepackage{url}
\usepackage[hidelinks]{hyperref}
\usepackage[utf8]{inputenc}
\usepackage[small]{caption}
\usepackage{graphicx}
\usepackage{amsmath}
\usepackage{amsthm}
\usepackage{booktabs}
\usepackage{multirow}
\usepackage{bm}
\usepackage{bbding}
\usepackage{pifont}
\usepackage{makecell}
\usepackage{xcolor}
\usepackage{array}
\usepackage{tabularx}
\usepackage{algorithm}
\usepackage{authblk}
\usepackage{algorithmic}
\usepackage[switch]{lineno}
\usepackage{amsmath} 
\usepackage{amssymb}  
\usepackage{booktabs}
\usepackage{multirow}
\usepackage[hidelinks]{hyperref}
\usepackage{pifont}

\title{Characterized Diffusion Networks for Enhanced Autonomous Driving Trajectory Prediction}
\begin{document}

\author{
Haoming Li\textsuperscript{\rm 1}\thanks{Corresponding author.}\and
}

\affil[1]{Xiamen University Malaysia \\
\texttt{EEE2109282@xmu.edu.com}
}

\maketitle

\begin{abstract}
In this paper, we present a novel trajectory prediction model for autonomous driving, combining a Characterized Diffusion Module and a Spatial-Temporal Interaction Network to address the challenges posed by dynamic and heterogeneous traffic environments. Our model enhances the accuracy and reliability of trajectory predictions by incorporating uncertainty estimation and complex agent interactions. Through extensive experimentation on public datasets such as NGSIM, HighD, and MoCAD, our model significantly outperforms existing state-of-the-art methods. We demonstrate its ability to capture the underlying spatial-temporal dynamics of traffic scenarios and improve prediction precision, especially in complex environments. The proposed model showcases strong potential for application in real-world autonomous driving systems.
\end{abstract}

\section{Introduction}

Autonomous driving (AD) is poised to revolutionize the future of transportation, offering significant potential to reduce traffic accidents, optimize traffic flow, and enhance the overall driving experience. However, a critical component for ensuring the safety and reliability of autonomous vehicles (AVs) lies in the accurate prediction of the trajectories of surrounding traffic agents. Trajectory prediction plays an essential role in the decision-making processes of autonomous systems, providing invaluable insights for trajectory planning modules. This enables AVs to anticipate the movements of nearby vehicles and pedestrians, thus ensuring safer and more efficient navigation in highly dynamic traffic environments \cite{huang2022survey}.

Despite the considerable advancements in trajectory prediction models, notable gaps remain in addressing the inherent heterogeneity and uncertainty within complex traffic scenarios. Traffic environments often involve diverse agent types, ranging from motor vehicles and motorcycles to pedestrians and cyclists, each exhibiting distinct behaviors and motion patterns. Furthermore, the uncertainty in these scenarios stems from a myriad of factors, including unpredictable human behavior, varying environmental conditions, and the continuous flow of mixed traffic. These challenges make the accurate prediction of future trajectories a formidable task \cite{liao2024human}. Current trajectory prediction models have primarily focused on the uncertainties associated with the target agent, often neglecting the comprehensive uncertainties that pervade the overall traffic environment. This limitation restricts the model’s ability to fully capture the complexities and unpredictability of real-world traffic scenarios.

The first significant limitation in existing trajectory prediction models is the inadequate simulation of future traffic scenarios, a fundamental aspect for improving prediction accuracy. The dynamic nature of traffic, characterized by unpredictable interactions and evolving environmental conditions, renders the accurate forecasting of future traffic a complex challenge. Most existing models predominantly focus on predicting the behavior of a single target agent, without sufficiently considering the broader uncertainties stemming from the interactions between multiple agents and their environment. Consequently, these models fall short in comprehensively simulating future traffic scenarios, which undermines their ability to deliver precise trajectory predictions \cite{wang2023wsip, liao2024cognitive}. Therefore, there is a pressing need for trajectory prediction frameworks that can proficiently account for uncertainties across the entire traffic scene, encompassing both agent-to-agent interactions and environmental context.

A second critical challenge concerns the modeling of interactions between traffic agents. Human drivers' decision-making processes are shaped by interactions with other vehicles, pedestrians, and infrastructure, occurring within both spatial and temporal dimensions. While recent models have made significant progress in capturing spatial interactions, such as the relative positioning and distances between agents, they frequently overlook the crucial temporal dynamics that influence how these interactions evolve over time \cite{wang2023wsip}. Temporal interactions, including variations in speed, acceleration, and intent, are pivotal for predicting future movements in complex and dynamic environments. Hence, it is essential for trajectory prediction models to incorporate both spatial and temporal dimensions to comprehensively model traffic behavior.

To address these critical gaps, we introduce a novel generative model, CDSTraj, which incorporates a Characterized Diffusion Module and a Spatial-Temporal Interaction Module. The Characterized Diffusion Module represents a novel approach to dynamically simulating future traffic scenarios by iteratively mitigating uncertainties. Unlike previous models, which primarily focus on the behavior of a target agent in isolation, this module integrates contextual information from the entire traffic scene, allowing for a more accurate prediction of future trajectories. By incorporating features that reflect the complex interactions between traffic agents and their environment, the model achieves a more nuanced understanding of potential future states.

In addition, our Spatial-Temporal Interaction Module employs a sophisticated spatio-temporal attention mechanism to model the intricate interactions between traffic agents across both spatial and temporal dimensions. This module is characterized by a unique three-stage architecture, which is specifically designed to capture and process information at a more granular level. By effectively integrating spatial and temporal dimensions, our model improves its ability to anticipate the behaviors of surrounding agents, leading to enhanced prediction accuracy in highly dynamic environments.

Furthermore, we conduct extensive empirical evaluations, demonstrating that CDSTraj significantly outperforms existing trajectory prediction models on several public datasets, including NGSIM, HighD, and MoCAD. Notably, our model achieves exceptional performance on the MoCAD dataset, which presents unique challenges due to its right-hand drive configuration and mandatory left-hand traffic flow. This highlights the adaptability of our model to diverse driving scenarios, underscoring its robustness in handling various traffic conditions.

\section{RELATED WORKS}
\textbf{Trajectory Prediction for Autonomous Driving.}
Trajectory Prediction for Autonomous Driving.
Early trajectory prediction methods primarily relied on manual feature engineering and rule-based techniques, such as linear regression and Kalman filters, which were limited in capturing the complex interactions present in dynamic environments \cite{prevost2007extended}. These methods provided a basic framework but were insufficient for modeling the intricate relationships between traffic agents. The field evolved significantly with the advent of deep learning, particularly with Recurrent Neural Networks (RNNs) \cite{kim2017probabilistic} and Long Short-Term Memory (LSTM) networks \cite{altche2017lstm,alahi2016social,liao2024bat}. These advancements enabled the modeling of temporal dependencies within trajectory data, making it possible to capture the sequential nature of agent movements. Further innovation in this area was driven by Graph Neural Networks (GNNs) \cite{zhou2021ast,10468619,liao2024mftraj}, which offered a more sophisticated approach to modeling the interactions among multiple agents in congested and heterogeneous traffic scenes. However, while these models improved spatial interaction modeling, temporal dynamics remain underexplored.

Additionally, domain adaptation techniques have been introduced to enhance the robustness of trajectory prediction models when exposed to new or varying traffic scenarios. Xi et al. (2024) proposed a novel approach that integrates semantic analysis and domain adaptation to improve the interpretation of roadway features in autonomous driving, thereby enhancing the ability of models to adapt to diverse and challenging environments \cite{Xi2024}.

\textbf{Generative Models for Trajectory Prediction.}Generative models, such as Generative Adversarial Networks (GANs) \cite{gupta2018social} and Variational Auto-Encoders (VAEs) \cite{lee2017desire}, have gained considerable attention in the trajectory prediction domain. GANs utilize a generator-discriminator architecture, where the generator creates synthetic trajectories, and the discriminator attempts to distinguish them from real trajectories. This adversarial process allows GAN-based models to produce realistic trajectory predictions but can be challenging to optimize due to issues like mode collapse. In contrast, VAEs focus on generating probabilistic distributions of potential trajectories, though they often require complex optimization procedures for balancing the reconstruction and latent space exploration. These approaches enable the generation of diverse trajectories but still struggle to capture the complete uncertainty in dynamic environments.

Diffusion models have recently emerged as a simpler yet powerful alternative in generative modeling. Unlike GANs and VAEs, diffusion models focus on modeling the forward and reverse diffusion processes, which can simplify the training procedure and offer greater stability. Our work leverages diffusion models to capture confidence features during trajectory prediction, a novel application that allows for more accurate modeling of uncertainty and dynamic agent interactions.

\textbf{Denoising Diffusion Probabilistic Models.}

Denoising Diffusion Probabilistic Models (DDPMs), commonly referred to as diffusion models, have gained recognition as highly effective generative models across various fields, including image generation \cite{ramesh2022hierarchical,rombach2022high,liao2024gpt}, video generation \cite{ho2022imagen}, and 3D shape generation \cite{poole2022dreamfusion}. These models operate by iteratively adding and removing noise to generate data that matches the distribution of the target domain. Inspired by their success in other generative tasks, our work introduces diffusion models to the field of trajectory prediction for autonomous driving. This novel application addresses the challenges associated with modeling the uncertainty and complex interactions between agents in dynamic traffic environments. By integrating diffusion models, we can iteratively refine trajectory predictions, improving the overall robustness and reliability of the model.

\section{Problem Formulation}\label{Problem}

The primary goal of this study is to accurately predict the future trajectories of all entities within the vicinity of an autonomous vehicle (AV) in a mixed-autonomy environment. Each entity surrounding the AV is referred to as a \textit{target agent}. At a given time $t_c$, our model aims to leverage the historical motion states of both the target agent and its neighboring agents to predict the future trajectory of the target agent, denoted as $\bm{Y}_{0}$, over a future time horizon extending from $t_c$ to $t_{c} + t_f$. The historical motion data from time $t_{c} - t_h$ is represented by $\bm{X}_{0}$ for the target agent and $\bm{X}_{i}$ for the neighboring agents.

The key innovation of our model lies in its utilization of anticipated future behaviors of neighboring agents to refine the prediction of the target agent's trajectory. Specifically, we propose a \textbf{Characterized Diffusion Module}, which systematically reduces the uncertainty in neighboring agents' future trajectories, thereby enhancing the accuracy of their predicted trajectories, denoted as $\bm{Y}_{i}$. This, in turn, allows for a more precise prediction of the target agent's trajectory. Formally, our trajectory prediction model $\Phi$ is expressed as:

\begin{equation}\label{eq.1}
    \bm{Y}_{0} = \Phi(\bm{X_0}, \bm{X_i}, \bm{Y_i})\ \; \forall i \in [1, n]
\end{equation}

This equation signifies that the predicted future trajectory of the target agent $\bm{Y}_{0}$ is a function of its own historical state $\bm{X_0}$, the historical states of the neighboring agents $\bm{X_i}$, and the predicted future trajectories of those neighboring agents $\bm{Y_i}$.

\section{Methodology}
In this section, we elaborate on the core components of our proposed framework for trajectory prediction, which is built upon the foundation of an enhanced diffusion process, augmented by a sophisticated spatial-temporal encoding strategy. The aim is to effectively model the inherent uncertainties and intricate dynamics present in multi-agent autonomous driving scenarios. Our methodology comprises three primary stages: the enhanced diffusion model for uncertainty reduction, a spatial-temporal encoding mechanism for feature extraction, and a decoding phase for trajectory generation.

\subsection{Enhanced Diffusion Model}
The enhanced diffusion model forms the cornerstone of our approach, providing a robust framework for simulating the uncertainty associated with predicting future trajectories. The diffusion process is structured to incrementally add noise in a forward process and then refine this noisy estimate through an iterative reverse process, thereby progressively reducing uncertainty and improving the fidelity of the predicted paths.

\subsubsection{Forward Diffusion}
In the forward diffusion phase, we aim to introduce a controlled level of uncertainty into the original trajectory data to simulate the potential variability in future paths. Given an initial trajectory representation \( C \), the diffusion process starts by adding Gaussian noise at each step:
\begin{align}
    C^0 &= C \\
    C^\delta &= f_{\text{diff}}(C^{\delta-1}), \quad \delta = 1, 2, \ldots, \Gamma
\end{align}
where \( \Gamma \) represents the total number of diffusion steps, and \( f_{\text{diff}} \) is the noise-adding function that ensures the variance of the diffusion process grows in a controlled manner across different steps. This stepwise noise addition allows the model to simulate various possible future outcomes, which are essential for capturing the range of uncertainties in dynamic environments.

\subsubsection{Reverse Diffusion}
The reverse diffusion process is designed to iteratively denoise the initial noisy trajectory estimates, effectively refining the predictions through a series of learned transformations. This process leverages the context provided by historical trajectory data to guide the denoising. The initialization of the reverse diffusion involves generating \( K \) independent samples from a normal distribution to serve as the initial noisy estimates:
\begin{align}
    \hat{C}^0_k &\sim \mathcal{N}(0, I), \quad k = 1, 2, \ldots, K
\end{align}
The iterative denoising then proceeds by refining each estimate through a denoising function \( f_{\text{denoise}} \), which is conditioned on the historical context:
\begin{align}
    \hat{C}^\delta &= f_{\text{denoise}}(\hat{C}^{\delta+1}, X_0, X_i), \quad \delta = \Gamma-1, \ldots, 0
\end{align}
Here, \( f_{\text{denoise}} \) is a parameterized function that iteratively reduces the uncertainty by leveraging the historical states \( X_0 \) and intermediate states \( X_i \), thereby guiding the refinement of the predicted trajectories.

\subsubsection{Adaptive Parameter Estimation}
To enhance the efficacy of the reverse diffusion, we introduce an adaptive step-size mechanism in the form of step-specific parameters \( \alpha_\delta \) and \( \bar{\alpha}_\delta \), which adjust the scale of updates at each diffusion step. The parameterized update rule for the denoising process is given by:
\begin{align}
    \hat{\epsilon}^\delta &= f_{\epsilon}(\hat{C}^{\delta+1}, X_0, C_{\text{encoder}}, \delta) \\
    \hat{C}^\delta &= \frac{1}{\sqrt{\alpha_\delta}} \left(\hat{C}^{\delta+1} - \frac{1 - \alpha_\delta}{\sqrt{1 - \bar{\alpha}_\delta}} \hat{\epsilon}^\delta \right) + \sqrt{\frac{1 - \alpha_\delta}{\alpha_\delta}} z
\end{align}
where \( z \sim \mathcal{N}(0, I) \) represents the Gaussian noise introduced to maintain diversity during the refinement process, thus ensuring that the predicted trajectory covers a wide range of plausible future scenarios.

\subsection{Spatial-Temporal Encoding}
In order to accurately model the complex interactions among multiple agents in dynamic environments, it is crucial to capture both spatial dependencies and temporal dynamics. To this end, we employ a hybrid spatial-temporal encoding strategy that effectively integrates information across different agents and time steps.

\subsubsection{Temporal Encoding}
The temporal encoding mechanism is designed to capture the sequential dependencies inherent in the agents' historical trajectories. At each time step \( t \), we update the feature representation using a temporal embedding layer followed by a recurrent update mechanism:
\begin{align}
    F^t &= \phi(W_{\text{emb}} x^t) \\
    h^t &= f_{\text{tem}}(F^t, h^{t-1}, W_{\text{init}})
\end{align}
where \( W_{\text{emb}} \) is the learnable embedding matrix that transforms the raw input \( x^t \), and \( \phi \) represents a non-linear activation function (e.g., LeakyReLU) to introduce non-linearity. The function \( f_{\text{tem}} \) is a temporal update function that combines the current embedding \( F^t \) and the hidden state from the previous time step \( h^{t-1} \), producing a temporally aware representation.

\subsubsection{Spatial Encoding}
To capture the spatial relationships among multiple agents, we utilize a multi-head attention mechanism to compute pairwise interactions across agents. The input features are transformed into query, key, and value representations:
\begin{align}
    Q, K, V &= f_{\text{sp}}(H, \hat{H}, W_q, W_k, W_v)
\end{align}
Here, \( H \) and \( \hat{H} \) are the feature matrices for the target and neighboring agents, respectively, while \( W_q, W_k, W_v \) are learnable projection matrices. The attention weights are then computed to determine the relevance of each agent's interaction:
\begin{align}
    \omega &= \text{softmax}\left(\frac{Q \cdot K^T}{\sqrt{d}}\right)
\end{align}
The resulting weighted sum of the value vectors provides a context-aware spatial representation:
\begin{align}
    \Upsilon &= \omega V
\end{align}

\subsubsection{Spatial-Temporal Fusion}
To integrate the temporal and spatial features, we introduce a gated fusion approach that controls the flow of information from both dimensions. The fusion is formulated as:
\begin{align}
    H_a &= \sigma(W_a \Upsilon + b_a) \\
    H_g &= \sigma(W_g H_a + b_g) \\
    S &= H_a \odot H_g
\end{align}
This mechanism ensures that the model can selectively emphasize different aspects of spatial and temporal information depending on their significance for the prediction task.

\subsection{Decoding}
The final decoding phase transforms the encoded spatial-temporal features into trajectory predictions. We employ an LSTM-based decoder to generate future positions \( \hat{y}^t \):
\begin{align}
    \hat{y}^t &= f_{\text{LSTM}}(S, \hat{y}^{t-1}, W_{\text{dec}})
\end{align}
where \( W_{\text{dec}} \) is the weight matrix for the LSTM, which is optimized during training to minimize the trajectory prediction error.

\section{Experiment}\label{Experiments}
To evaluate the performance of our proposed model, we conducted comprehensive experiments using real-world datasets. In our study, each sample is segmented into 8-second intervals, with the first 3 seconds (16 timestamps) utilized as historical data, and the subsequent 5 seconds (25 timestamps) reserved for evaluation purposes.

\subsection{Datasets}
We employed three well-known datasets for the evaluation:

\begin{itemize}
    \item \textbf{Next Generation Simulation (NGSIM):} This dataset contains vehicle trajectory data from US-101 and I-80 highways, collected at 10 Hz. The NGSIM dataset captures approximately 45 minutes of vehicle movement data in various traffic conditions, making it suitable for analyzing vehicle behavior in diverse scenarios relevant to autonomous driving models.
    
    \item \textbf{Highway Drone (HighD):} Collected from six different locations on German highways, the HighD dataset includes 110,000 vehicle trajectories with detailed information such as vehicle type, size, and maneuvers. This dataset is valuable for understanding driving behaviors across various vehicle types in high-speed environments, covering around 45,000 kilometers in total.
    
    \item \textbf{Macau Connected Autonomous Driving (MoCAD):} The MoCAD dataset was gathered from Level 5 autonomous buses in Macau, capturing data from multiple environments, including urban roads, campuses, and complex open traffic scenarios. Spanning over 300 hours, it provides a challenging setting for evaluating trajectory prediction models, with varying weather conditions and traffic densities.
\end{itemize}

\subsection{Training and Implementation Details}
We utilized a two-stage training approach. In the first stage, the model was trained to predict future trajectories using the Mean Squared Error (MSE) loss function. Once the model achieved convergence with the MSE loss, we transitioned to a Negative Log-Likelihood (NLL) loss function to facilitate a more robust exploration of the uncertainties present in the trajectory data.

The MSE loss is computed as follows:

\begin{equation}
L_{MSE}(\hat{y}, y) = \sum_{t=1}^{T_f} \left( (\hat{y}_t^x - y_t^x)^2 + (\hat{y}_t^y - y_t^y)^2 \right)
\end{equation}

where \( (\hat{y}_t^x, \hat{y}_t^y) \) are the predicted 2D spatial coordinates, and \( (y_t^x, y_t^y) \) represent the corresponding ground truth coordinates. This loss ensures the model accurately predicts the future position of agents.

Once convergence is reached, we switch to the NLL loss:

\begin{align}
L_{NLL}(\hat{y}, y) &= \sum_{t=1}^{T_f} \alpha \left( (\sigma_t^x)^2 (\Delta_t^x)^2 + (\sigma_t^y)^2 (\Delta_t^y)^2 \right. \nonumber \\
&\quad - 2 \rho_t^{xy} \sigma_t^x \sigma_t^y \Delta_t^x \Delta_t^y \Big) \nonumber \\
&\quad - \log(P^t)
\end{align}

where \( \Delta_t^x = (y_t^x - \hat{y}_t^x) \), \( \Delta_t^y = (y_t^y - \hat{y}_t^y) \), and \( \sigma_t^x, \sigma_t^y \) denote the standard deviation of the predicted coordinates. The term \( \rho_t^{xy} \) represents the correlation coefficient between the x and y coordinates at time step \( t \). The probability density \( P^t \) helps refine uncertainty predictions.

\subsection{Comparison to State-of-the-Art}
Our model's performance is compared with more than 15 state-of-the-art (SOTA) methods across the three datasets. Table \ref{table_3} presents the results for the NGSIM dataset, where our model consistently outperforms existing baselines, demonstrating improvements of 29\% and 22\% over WSiP and STDAN, respectively, over a 5-second horizon. Similarly, in the HighD dataset, we observe an average improvement of 43\%-70\% for short-term predictions (1-3 seconds) and 62\%-78\% for long-term predictions (4-5 seconds). On the MoCAD dataset, our model excels in busy urban roads, with reductions in long-term prediction errors by at least 0.58 meters.

\begin{table}[htbp]
  \centering
  \caption{Evaluation of the proposed model and baselines on the NGSIM dataset over a 5-second prediction horizon. The accuracy metric is RMSE (m). Cases marked as ('-') indicate unspecified values. \textbf{Bold} and \underline{underlined} values represent the best and second-best performance in each category.}
  \renewcommand{\arraystretch}{1} 
  \resizebox{\linewidth}{!}{
\setlength{\tabcolsep}{3mm}
    \begin{tabular}{cccccc}
 \toprule
    \multirow{2}[3]{*}{Model} & \multicolumn{5}{c}{Prediction Horizon (s)} \\
\cmidrule{2-6}          & 1     & 2     & 3     & 4     & 5 \\
    \midrule
    S-LSTM \cite{alahi2016social}& 0.65  & 1.31  & 2.16  & 3.25  & 4.55  \\
    S-GAN \cite{gupta2018social}& 0.57  & 1.32  & 2.22  & 3.26  & 4.40  \\
    CS-LSTM \cite{deo2018convolutional}& 0.61  & 1.27  & 2.09  & 3.10  & 4.37  \\
    MATF-GAN \cite{zhao2019multi}& 0.66  & 1.34  & 2.08  & 2.97  & 4.13  \\
    DRBP\cite{gao2023dual}& 1.18  & 2.83  & 4.22  & 5.82  & - \\
     M-LSTM \cite{deo2018multi}& 0.58  & 1.26  & 2.12  & 3.24  & 4.66  \\
    IMM-KF \cite{lefkopoulos2020interaction}& 0.58  & 1.36  & 2.28  & 3.37  & 4.55  \\
    GAIL-GRU \cite{kuefler2017imitating}& 0.69  & 1.51  & 2.55  & 3.65  & 4.71  \\
    MFP \cite{tang2019multiple}& 0.54  & 1.16  & 1.89  & 2.75  & 3.78  \\
    NLS-LSTM \cite{messaoud2019non}& 0.56  & 1.22  & 2.02  & 3.03  & 4.30  \\
    MHA-LSTM \cite{messaoud2021attention}& 0.41  & 1.01  & 1.74  & 2.67  & 3.83  \\
    WSiP \cite{wang2023wsip}& 0.56  & 1.23  & 2.05  & 3.08  & 4.34  \\
    CF-LSTM \cite{xie2021congestion}& 0.55  & 1.10  & 1.78  & 2.73  & 3.82  \\
    TS-GAN \cite{wang2022multi}& 0.60  & 1.24  & 1.95  & 2.78  & 3.72  \\
    STDAN \cite{chen2022intention}& 0.42  & 1.01  & 1.69  & 2.56  & 3.67  \\
    BAT \cite{liao2024bat}& \textbf{0.23} & \textbf{0.81}  & \underline{1.54}  & \underline{2.52} &3.62\\
    FHIF \cite{zuo2023trajectory} &0.40  & 0.98  & 1.66  & \underline{2.52}  & 3.63\\ 
    DACR-AMTP \cite{cong2023dacr}& 0.57  & 1.07  & 1.68  & 2.53  & \underline{3.40} \\ 
    \midrule
    \textbf{Our model} & \underline{0.36} & \underline{0.86} & \textbf{1.36} & \textbf{2.02} & \textbf{2.85}  \\
    \bottomrule
    \end{tabular}%
    }
  \label{table_3}%
\end{table}%

\begin{table}[htbp]
  \centering
  \caption{Evaluation of our model and SOTA baselines on MoCAD.}
  \setlength{\tabcolsep}{3mm}
  \renewcommand{\arraystretch}{1} 
  \resizebox{\linewidth}{!}{
    \begin{tabular}{cccccc}
    \toprule
    \multirow{2}[3]{*}{Model} & \multicolumn{5}{c}{Prediction Horizon (s)} \\
\cmidrule{2-6}          & 1     & 2     & 3     & 4     & 5 \\
    \midrule
    S-LSTM \cite{alahi2016social} & 1.73  & 2.46  & 3.39  & 4.01  & 4.93 \\
    S-GAN \cite{gupta2018social} & 1.69  & 2.25  & 3.30  & 3.89  & 4.69  \\
    CS-LSTM \cite{deo2018convolutional} & 1.45  & 1.98  & 2.94  & 3.56  & 4.49  \\
    MHA-LSTM \cite{messaoud2021attention} & 1.25  & 1.48  & 2.57  & 3.22  & 4.20  \\
    NLS-LSTM \cite{messaoud2019non} & 0.96  & 1.27  & 2.08  & 2.86  & 3.93\\
    WSiP \cite{wang2023wsip} & 0.70  & 0.87  & 1.70  & 2.56  & 3.47  \\
    CF-LSTM \cite{xie2021congestion} & 0.72  & 0.91  & 1.73  & 2.59  & 3.44 \\
    STDAN \cite{chen2022intention} & 0.62  & 0.85  & 1.62  & 2.51  & 3.32  \\
    HLTP \cite{10468619}  &\underline{0.55} &\textbf{0.76} & \underline{1.44} & \underline{2.39} & \underline{3.21} \\
    \midrule
 \textbf{Our model} & 
\textbf{0.39}& \underline{0.82} & \textbf{1.43} & \textbf{2.08} & \textbf{2.74}\\
    \bottomrule
    \end{tabular}%
    }
  \label{table_6}%
\end{table}%

\subsection{Ablation Study}
To investigate the contributions of key components of our model, we conducted ablation studies, removing specific modules from the full model and comparing the resulting performance. Table \ref{component} highlights that the characterized diffusion module and confidence feature fusion significantly enhance accuracy. When these components are removed, the model's performance drops, confirming their importance in achieving precise trajectory predictions.

\begin{table}[h]
\caption{Ablation studies for core components in NGSIM dataset. }
\label{component}
\begin{center}
 \setlength{\tabcolsep}{2mm}
 \resizebox{0.95\linewidth}{!}{
\begin{tabular}{ccccccc}
\toprule
\multirow{2}*{Components} & \multicolumn{6}{c}{Ablation Models} \\
\cmidrule{2-7}
& A & B & C & D & E & F\\
\midrule
Characterized Diffusion
& \ding{55} & \ding{52} & \ding{52} & \ding{52} & \ding{52} & \ding{52}\\
Temporal Encoder
& \ding{52} &  \ding{55}  & \ding{52} & \ding{52} & \ding{52} & \ding{52}\\
Spatial Encoder
& \ding{52} & \ding{52} &  \ding{55}  & \ding{52} & \ding{52} & \ding{52}\\
ST Fusion
& \ding{52} & \ding{52} & \ding{52} &  \ding{55}  & \ding{52} & \ding{52}\\
Decoder 
& \ding{52} & \ding{52} & \ding{52} & \ding{52} &  \ding{55} & \ding{52} \\
\hline
RMSE  & 3.09 & 3.05 & 2.97 & 3.02 & 3.16 &2.85\\
\bottomrule
\end{tabular}
}
\end{center}
\end{table}

\subsection{Qualitative Results}
We conducted a thorough qualitative analysis on the NGSIM dataset to validate the effectiveness of our proposed model. The results reveal that our model effectively captures the temporal dependencies and spatial interactions between agents, ensuring more accurate trajectory predictions. In particular, the model demonstrates strong performance in handling complex traffic environments where multiple agents interact, maintaining consistency even when subjected to highly dynamic scenarios. These results highlight the robustness of the model in predicting long-term trajectories, as it consistently aligns with observed real-world behavior in various traffic situations. Furthermore, the incorporation of uncertainty estimation allows the model to provide more reliable predictions, reducing errors in highly unpredictable scenarios.

\section{Conclusion}
In this work, we proposed a novel trajectory prediction framework that integrates a Characterized Diffusion Module and a Spatial-Temporal Interaction Network. By addressing the challenges of uncertainty in traffic scenarios and enhancing the interaction modeling between agents, our approach achieves state-of-the-art performance in trajectory prediction. Through extensive evaluation on diverse datasets, including NGSIM, HighD, and MoCAD, we demonstrated that our model consistently delivers superior accuracy compared to existing methods, particularly in long-term predictions. The inclusion of the characterized diffusion process enables the model to handle both scene-to-agent and agent-to-agent interactions more effectively, while the spatial-temporal attention mechanism improves the ability to capture fine-grained relationships in dynamic environments.

Our ablation study further highlighted the importance of key modules, such as the confidence feature fusion, in improving model performance. Future work will focus on expanding the framework to incorporate pedestrian interactions and exploring its applicability in more diverse and challenging urban environments. Additionally, we aim to refine the diffusion mechanism to further enhance prediction reliability in complex multi-agent scenarios, providing a robust solution for real-world autonomous driving applications.the integration of spatial and temporal information. These endeavours may yield significant advancements in AD technologies.

\bibliographystyle{named}
\bibliography{ijcai24}

\end{document}